\documentclass{article}

\usepackage[numbers]{natbib}
\usepackage{tcolorbox}

\usepackage{wrapfig}
    \usepackage[preprint]{neurips_2023}

\usepackage[T1]{fontenc}    % use 8-bit T1 fonts
\usepackage{hyperref}       % hyperlinks
\usepackage{url}            % simple URL typesetting
\usepackage{booktabs}       % professional-quality tables
\usepackage{amsfonts}       % blackboard math symbols
\usepackage{nicefrac}       % compact symbols for 1/2, etc.
\usepackage{microtype}      % microtypography
\usepackage[normalem]{ulem}
\usepackage{xcolor}         % colors
\usepackage{adjustbox}
\usepackage{multirow}
\usepackage{amssymb}
\usepackage{amsmath}
\usepackage{pifont}
\newcommand{\cmark}{\ding{51}}%
\newcommand{\xmark}{\ding{55}}%

\usepackage{CJK}

\title{M$^3$IT: A Large-Scale Dataset towards \\ Multi-Modal Multilingual Instruction Tuning}

\newcommand*{\affaddr}[1]{#1} 
\newcommand*{\affmark}[1][*]{\textsuperscript{#1}}

\author{%
  Lei Li\affmark[$\dagger$], 
  Yuwei Yin\affmark[$\dagger$],
  Shicheng Li\affmark[\S],
  Liang Chen\affmark[\S],
  Peiyi Wang\affmark[\S],
  Shuhuai Ren\affmark[\S],
  Mukai Li\affmark[\ddag] \\ 
  \textbf{
  Yazheng Yang\affmark[$\dagger$],
  Jingjing Xu\affmark[\ddag],
  Xu Sun\affmark[\S],
  Lingpeng Kong\affmark[$\dagger$],
  Qi Liu\affmark[$\dagger$]} \\
  \affaddr{\affmark[$\dagger$] The University of Hong Kong}\\
  \affaddr{\affmark[\S] National Key Laboratory for Multimedia Information Processing,\\
      School of Computer Science, Peking University}\\
  \affaddr{\affmark[\ddag] Shanghai AI Lab}\\
  \texttt{nlp.lilei@gmail.com} \\ 
  \texttt{jingjingxu@pku.edu.cn} \quad \texttt{\{lpk, liuqi\}@cs.hku.hk}\\
}

\begin{document}

\maketitle

\begin{abstract}
% \qi{add one sentence comparing our dataset with respect to previous ones, e.g. more large-scale, more natural responses etc.} 
% Instruction tuning has emerged as a powerful approach for aligning large language models (LLMs) to human instructions across various tasks. Leveraging large-scale instruction data greatly contributes to building 
% phenomenal generative AI systems such as ChatGPT.
% However, beyond text-only models, progress in open vision-language models (VLMs) has been constrained by the lack of high-quality instruction datasets.
% To address this issue and democratize research on large-scale general-purpose
% models, we present a multi-modal, multilingual instruction tuning~(M$^3$IT) dataset to enhance the alignment of VLMs with human instructions. We carefully curate $40$ datasets, including $2$ million instances with $400$ manually written task instructions, into a multimodal-to-text dialog format, with key tasks translated into $101$ languages.
% Our M$^3$IT outperforms previous datasets regarding the scale of covered tasks, task instructions and instances, and response quality.
% Furthermore, we build Ying-VLM, a VLM model trained on our dataset, which demonstrates great potential to answer challenging answers that require world knowledge, generalize to unseen video tasks and understand unseen instructions in Chinese. 
% We have open-sourced the dataset and trained models to facilitate future research.

Instruction tuning has significantly advanced large language models (LLMs) such as ChatGPT, enabling them to align with human instructions across diverse tasks. However, progress in open vision-language models (VLMs) has been limited due to the scarcity of high-quality instruction datasets. To tackle this challenge and promote research in the vision-language field, we introduce the Multi-Modal, Multilingual Instruction Tuning (M$^3$IT) dataset, designed to optimize VLM alignment with human instructions.
Our M$^3$IT dataset comprises 40 carefully curated datasets, including 2.4 million instances and 400 manually written task instructions, reformatted into a vision-to-text structure. Key tasks are translated into 80 languages with an advanced translation system, ensuring broader accessibility. M$^3$IT surpasses previous datasets regarding task coverage, instruction number and instance scale.
Moreover, we develop Ying-VLM, a VLM model trained on our M$^3$IT dataset, showcasing its potential to answer complex questions requiring world knowledge, generalize to unseen video tasks, and comprehend unseen instructions in Chinese. We have open-sourced the dataset to encourage further research.\footnote{Our dataset is available at \url{https://huggingface.co/datasets/MMInstruction/M3IT}}
% [TODO]
%We curate 40 datasets and manually create tailored instructions, while also refining visual and textual inputs for improved naturalness. The resulting M2BI dataset comprises 400 instructions and instances over 2 million.
%%%%
%Our methodology involves unifying existing multi-modal tasks into a multimodal-to-text format like GPT-4. , encompassing classification, captioning, generation, reasoning, and question-answering. 
%We then curate diverse datasets for each task type and manually create tailored instructions, while also refining visual and textual inputs for improved naturalness. The resulting M2BI dataset comprises 400 instructions and instances over 2 million.
%We investigate cross-task generalization using this dataset across three dimensions: difficulty, language (English to Chinese), and modality (Images to Videos). 
%Experimental results reveal that instruction tuning not only enables large language models to tackle complex reasoning tasks but also effectively generalizes across languages and modalities.
%Further analysis underscores the importance of instruction diversity, task variety, and data-balancing strategies in enhancing the model's ability to follow human instructions accurately. 

\end{abstract}

\section{Introduction}

% Background
There has been a continuously increasing trend to develop intelligent assistants that can follow human instructions~\citep{gpt3,chatgpt,openai2023gpt4}. 
In the natural language processing (NLP) field, instruction tuning~\citep{mishra2022naturalinstruction, weifinetuned} is a success paradigm that leverages large-scale well-formatted instances to align large language models (LLMs) to human instructions. 
By finetuning on instances with specific task descriptions, LLMs learn to follow the instruction to perform various tasks, and demonstrate strong generalization ability on unseen tasks
~\citep{longpre2023flan}.
Expanding beyond NLP, a general-purpose intelligent agent must encompass various modalities, such as vision, prompting recent efforts to investigate instruction tuning in vision-language domains~\citep{zhu2023minigpt4,liu2023llava,dai2023instructblip}. 
To develop powerful vision-language models (VLMs), it is essential to have a well-constructed dataset that encompasses diverse vision-language tasks and aligns with human instructions.
However, the instructional data supporting existing VLMs is either not publicly available (e.g., GPT-4) or offers limited task and language coverage (e.g., only tasks in English are considered). This scarcity of comprehensive datasets has impeded the progress of open vision-language models, highlighting the importance of multi-modal instruction tuning and the need for high-quality datasets.
In this paper, we aim to advance instruction tuning research in the multi-modal domain by introducing an open dataset M$^3$IT, a \textbf{M}ulti-\textbf{M}odal \textbf{M}ultilingual \textbf{I}nstruction \textbf{T}uning dataset, as an essential step towards building a versatile general-purpose assistant. 
We build this dataset by converting existing datasets into a unified vision-to-text schema with four stages: (1) manual instruction writing, (2) dataset pre-processing, (3) careful quality check and (4) dataset translation for key tasks. 
%We achieve this by converting existing high-quality multi-modal datasets into a unified schema through a three-step process: (1) enlisting multi-modal experts to understand the purpose of each dataset/project and compose diverse instructions, (2) transforming the original dataset by processing images for region-based tasks and paraphrasing brief answers with the powerful GPT-4 model, supplemented with additional context, and (3) conducting a thorough manual quality check by randomly sampling $10$ instances from each dataset.
Our dataset encompasses a wide range of tasks, including classic image-text tasks such as image classification, visual question answering, and image captioning. 
Video-related tasks, such as video question-answering, are also incorporated to ensure comprehensive coverage across multiple modalities.
We further integrate Chinese vision-language datasets with corresponding Chinese instructions.
The resulting dataset compiles 40 diverse tasks and 400 instructions.
Finally, key vision-language tasks are translated into 80 languages with a strong translation system, to support multilingual studies.

% Experiments
To evaluate the effectiveness of the proposed dataset, we develop a vision-language model, Ying-VLM, by integrating a strong vision encoder, BLIP-2~\citep{li2023blip2} with a large language model, Ziya-13B~\citep{fengshenbang}, derived from LLaMA~\citep{touvron2023llama}. 
Building on the successful approach of incorporating visual tokens as textual prompts in LLMs~\citep{dai2023instructblip,zhu2023minigpt4,liu2023llava}, we employ a two-stage training process: (1) the initial stage aligns vision features with text embeddings through image captioning on LAION400M~\citep{laion400m}, and (2) the second stage enhances the model by conducting instruction tuning on selected tasks of our dataset. %with tasks held-out for evaluation.
Experimental results reveal that Ying-VLM surpasses strong baseline models in knowledgeable VQA tasks and exhibits improved generalization performance to unseen video and cross-lingual tasks.
Further analysis indicates that the improved performance corresponds to increased tasks for instruction tuning, while the diversity of instructions also affects outcomes.

% Contribution
% The contribution of this paper is two-fold:
% (1) We open-source a large-scale multi-modal multilingual instruction tuning~(M$^3$IT) dataset to facilitate general-purpose multi-modal agents.
% (2) We build Ying-VLM, a visual assistant that performs well on knowledgeable VQA and generalizes well to unseen video QA and Chinese multi-modal tasks, providing useful insights for future studies towards building general-purpose intelligent agents.
This paper presents two key contributions:
(1) We introduce the open-source, large-scale Multi-modal, multilingual Instruction Tuning (M$^3$IT) dataset, designed to enable the development of general-purpose multi-modal agents.
(2) We develop Ying-VLM, a visual assistant that excels in knowledgeable VQA tasks, demonstrates strong generalization to unseen video QA and Chinese multi-modal tasks, and offers valuable insights for future research.

\section{Related Work}

\begin{table}[ht!]
    \centering
    
    \caption{Summary of multi-modal instruction tuning datasets. }
    \small 
    \begin{adjustbox}{max width=\textwidth}
    \begin{tabular}{@{}lccccc@{}}
    \toprule
       Dataset  &  \# Tasks & Multi-Lingual &  \# of Instances & Avg. \# of Manual Instructions / Task & Open-Sourced\\
       \midrule
        MiniGPT4 & N / A & \xmark & 5K & N / A & \cmark\\ 
        LLaVA & 3  & \xmark & 1.15M& N / A &  \cmark \\ 
         MultiModalGPT &  3 & \xmark & 6K &  5 & \xmark \\ 
       MultiInstruct &  26 & \xmark & $\sim$ 235K  &  5 & \xmark \\ 
      
       InstructBLIP & 28 & \xmark & $\sim$ 1.6M & 9.7 & \xmark \\ 
        M$^3$IT~(Ours) & 40 & \cmark &  2.4M  & 10   & \cmark\\ 
    \bottomrule
    \end{tabular}
    \end{adjustbox}
    \label{tab:dataset_compare}
\end{table}

% Our work is inspired by recent progress in natural language instruction tuning and multi-modal instruction tuning.

%\paragraph{Natural Language Instruction Tuning.}
Our work draws inspiration from recent language instruction tuning benchmarks~\citep{weifinetuned,mishra2022naturalinstruction}, which have been proven effective for improving language models to obtain cross-task generalization ability~\citep{longpre2023flan,wang2022superNI}. 
In this paper, we focus on exploring the instruction tuning paradigm from LLMs to multi-modal agents. Unlike text-only tasks, vision-language tasks generally have more diverse formats, which poses new challenges toward vision-language instruction tuning benchmarks.

To develop a general-purpose vision-language model, it is crucial to create high-quality multi-modal instruction tuning datasets encompassing diverse tasks, languages, and instructions.
Several studies have investigated multi-modal instruction tuning for VLMs. 
LLaVA~\citep{liu2023llava} and MiniGPT-4~\citep{zhu2023minigpt4} generate visual content-related dialog by incorporating image caption data into GPT-4/ChatGPT models.
MultiInstruct~\citep{xu2022multiinstruct} reformats a series of visual classification tasks into an instruction-tuning format, while InstructBLIP~\citep{dai2023instructblip} adapts 28 existing image-to-text tasks.
However, these datasets do not provide an ideal multi-modal instruction tuning dataset due to their limited (1) coverage of various task types in multi-modal fields, (2) diversity and quality of instances, and (3) inclusion of multiple languages for wide linguistic diversity. In this paper, we construct an improved multi-modal instruction tuning dataset by expanding task coverage to 40 datasets, supplementing instances with 10 manually written task instructions, and including tasks in different languages.
Table~\ref{tab:dataset_compare} compares the characteristics of existing multi-modal instruction tuning datasets and M$^3$IT.

\section{M$^3$IT: A Multi-Modal Multilingual Instruction Tuning Dataset}
In this section, we introduce our proposed M$^3$IT dataset by first elaborating the dataset coverage~(\S~\ref{subsec:task_coverage}), followed by the details of the annotation process(\S~\ref{subsec:annotation_process}).
Finally, we present the dataset format and provide the statistics of the crafted datasets instructions(\S~\ref{subsec:data_format}).

% We should highlight the difference with the previous benchmark.
% rewording answers, Extend to Videos, multilingual 

\subsection{Task Coverage}
\label{subsec:task_coverage}
Our dataset compiles diverse tasks of classical vision-language tasks, including captioning, visual question answering~(VQA), visual conditioned generation, reasoning and classification.

\noindent\textbf{Captioning} This task aims to produce descriptions of the given images according to different needs. We include
MS COCO~\citep{lin2014mscoco} (the Karpathy split) for generic image descriptions.
TextCaps~\citep{sidorov2020textcaps} requires models to capture the text presented in the image and generate captions accordingly.  
Image-Paragraph-Captioning~\citep{krause2016image_para_cap} focuses on generating detailed descriptions for images.
% textcap
% COCO-CN~\citep{li2019coco_cn},  % coco-cn
 % image-paragraph-captioning
% nocaps~\citep{agrawal2019nocaps},  % nocaps
% Visual Genome~\citep{krishna2017visual_genome},  % Visual Genome
% MSR-VTT~\citep{xu2016msrvtt},  % msrvtt
% Flickr8K-CN~\citep{li2016flickr8k_cn},  % flickr8k-cn

\noindent\textbf{Reasoning} This task evaluates specific reasoning capabilities. We incorporate
CLEVR~\citep{johnson2017clevr} and  % clevr
NLVR~\citep{Suhr2017NLVR} for spatial reasoning,  % nlvr
Visual Commonsense Reasoning (VCR)~\citep{zellers2019vcr} for commonsense reasoning,  % vcr
Visual MRC~\citep{tanaka2021visualmrc} for reading comprehensive over images,  % visual-mrc
and Winoground~\citep{thrush2022winoground} for fine-grained semantics reasoning over text descriptions and image contents.  % winoground

\noindent\textbf{Visual Question Answering (VQA)} 
This is the most widely studied multi-modal task, which requires the model to answer a given question based on the image correctly. Tasks include
VQA v2~\citep{balanced_vqa_v2},  % vqa-v2
Shapes VQA~\citep{andreas2016shapes},  % shapes
DocVQA~\citep{mathew2021docvqa},  % docvqa
OCR-VQA~\citep{mishra2019ocr_vqa},  % ocr-vqa
ST-VQA~\citep{biten2019st_vqa},  % st-vqa
Text-VQA~\citep{singh2019text_vqa},  % text-vqa
and GQA~\citep{hudson2019gqa}.  % gqa

\noindent\textbf{Knowledgeable Visual Question Answering}
Unlike traditional VQA tasks focusing on the question relevant to the content image, knowledgeable visual question answer~(KVQA) requires the model to draw upon outside knowledge to answer questions. We incorporate two outside knowledge VQA datasets:
OK-VQA~\citep{marino2019okvqa} and A-OK-VQA~\citep{schwenk2022aokvqa},  % a-okvqa
ScienceQA~\citep{lu2022scienceqa} which 
contains multi-modal science questions,  % science-qa
and ViQuAE~\citep{lerner2022viquae} focusing on knowledge facts of named entities in images.  % viquae

\noindent\textbf{Classification}
This task involves classifying an image based on a given set of candidate labels.
ImageNet~\citep{russakovsky2015imagenet},
Grounded Object Identification (COCO-GOI)~\citep{lin2014mscoco},  % coco-goi
COCO-Text~\citep{veit2016coco_text},  % 
Image Text Matching (COCO-ITM)~\citep{lin2014mscoco},  % coco-itm 
e-SNLI-VE~\citep{esnlive},  % snli-ve
Multi-modal Fact Checking (Mocheg)~\citep{yao2022mocheg},
and IQA~\citep{duanmu2021iqa} are included.  % iqa
Due to language model input length constraints, we reduce the number of options in some datasets with extensive candidate labels, such as ImageNet.
% Due to the length constraint of language model inputs, we reduce the number of options the on some datasets with huge candidate labels, such as ImageNet. \qi{much typos here}

\noindent\textbf{Generation}
Visual conditional general requires models to understand the visual content and make a composition meeting the task demand. We have
Visual Storytelling (VIST)~\citep{huang2016vist},  % vist
Visual Dialog (VisDial)~\citep{das2017visual_dialog},  % visual-dialog
and multi-modal machine translation
Multi30k~\citep{elliott2016multi30k} in this category.  % multi30k
%,  % mmchat

\noindent\textbf{Chinese and multilingual Vision-Language Tasks}
To examine the effect of instruction tuning on different languages, we incorporate several Chinese vision-language tasks including FM-IQA~\citep{fm-iqa} for VQA, COCO-CN~\citep{li2019cococn} and Flickr8k-CN ~\citep{flickr8kcn} for captioning, Chinese Food Net~\citep{chen2017chinesefoodnet} for classification, and MMChat~\citep{zheng2021mmchat} for generation.

\noindent\textbf{Video-Language Tasks} 
Beyond the static images, we are interested in whether instruction tuning can also be applied to video-text tasks. We include the classic MSR-VTT datasets~\citep{xu2016msrvtt} for video captioning,  MSRVTT-QA~\citep{xu2017msrvtt_qa}, ActivityNet-QA~\citep{yu2019activitynet}, iVQA~\citep{yang2021iVQA} and MSVD-QA~\citep{xu2017msrvtt_qa} for video question answering, Something-Something~\citep{goyal2017something_something} for video action classification.
% MSR-VTT-QA~\citep{xu2017msrvtt_qa},  % msrvtt-qa
% To add: activitynet-qa, ivqa, msvd-qa

\begin{figure}[t!]
    \centering
    \includegraphics[width=\linewidth]{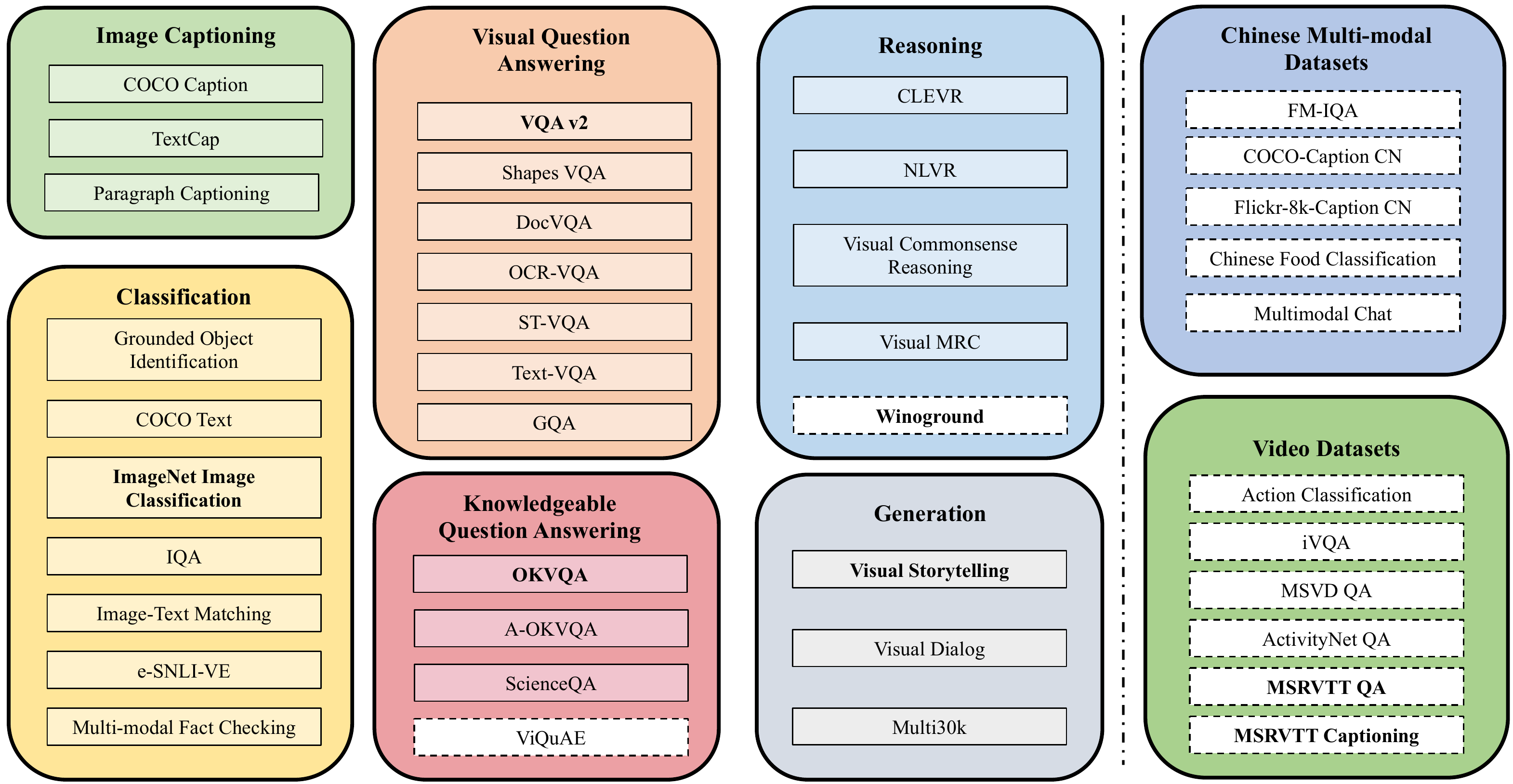}
    \caption{Tasks in our proposed multi-modal multilingual instruction tuning dataset. The tasks in dashed white boxes are held-out evaluation sets that are not adopted during training. Tasks with bold names are translated into $80$ languages.}
    \label{fig:dataset_tasks}
\end{figure}

As shown in Figure~\ref{fig:dataset_tasks}, our dataset makes a wide coverage of the current existing visual-language and video-language benchmarks, enabling different skill sets for the language models, from simple image captioning to complicated reasoning based on the image even beyond the visual content.

\subsection{Annotation Process}
\label{subsec:annotation_process}

\begin{table}[htb!]
    \centering
    \caption{The statistics of our instructions.}
    \begin{adjustbox}{max width=0.8\textwidth}
    \begin{tabular}{lr}
    \toprule
    Number of different instructions & 400 \\
    \quad - Image Captioning & 52 \\
    \quad - Classification & 113 \\
    \quad - Visual Question Answering & 95 \\
    \quad - Knowledgeable Visual QA & 40 \\
    \quad - Reasoning & 60 \\
    \quad - Generation & 40 \\
    \midrule
    Tokens per instruction & $24.4\pm9.6$  \\
    \midrule
    Instruction edit distance among the same task & $76.6\pm37.2$ \\
    \midrule
    Instruction edit distance across tasks & $106.6\pm39.5$ \\
    \bottomrule
    \end{tabular}
    \end{adjustbox}
    \label{tab:instruct_stat}
\end{table}

To build high-quality multi-modal instruction datasets, we rewrite various datasets into a vision-to-text format. The annotation process includes four steps: (1) writing instructions for each task, (2) structuring images and texts into a unified schema, (3) checking the overall dataset quality, and (4) building multilingual sets. 
Eight authors of this work are employed as human annotators, each of whom is a graduate student familiar with relevant literature.

\noindent\textbf{Stage I: Instruction Writing}
To build high-quality instructions, we first ask annotators to carefully read the dataset paper and check the original dataset with some instances to get a clear understanding of the task. After that, they are required to write $10$ diverse task instructions manually, covering the key characteristics of the task.
%vary the expression so that the instructions can be diverse in wording and  syntactic structures
%Besides, they are guided to write diverse instructions. 
Table~\ref{tab:instruct_stat} shows the statistics of the written instructions for each task. In total, we annotate $400$ instructions for all tasks. The average length per instruction is $24.4$. To evaluate the diversity of annotated instructions, we employ the average edit distance to measure the similarity between two strings. The average edit distance within the same task is $76.6$, indicating a good range of instruction diversity.

\noindent\textbf{Stage II: Data Format Unification}
After the instruction has been written according to the task characteristics, we further process the images and corresponding text for a unified instance schema.
For most datasets, we keep the original images and text, where images are converted into corresponding base64 encoded strings for easy data loading. 
We perform two modifications on potential examples: (1)
\noindent\textbf{Adding Bounding Box to Images.}
For tasks designed for specific regions in the image, a straightforward solution is to provide the bounding box information in natural language for informing the language models of the regions in interest. 
However, the image preprocessing techniques adopted by different vision encoders may resize the original image, and the original bounding box annotation thus needs further adjustments. 
Inspired by the recent observation that common vision encoders such as CLIP~\citep{radford2021clip} are sensitive to the visual prompt~\citep{shtedritski2023does}, we directly tag the bounding box as a red rectangle to the image, serving as a hint for VLMs to focus on the target region.
% Another option is to introduce extra box tokens into the language model's vocabulary, which requires heavy retraining of the language model for a better word representation of these newly introduced tokens.
(2)
\noindent\textbf{Short Answer Paraphrasing.}
As recent studies have shown that the original short and brief answers in the common VQA dataset could negatively influence the model generation performance~\citep{dai2023instructblip}, 
we propose to utilize the ChatGPT~\citep{chatgpt} model for paraphrasing the original answers, by providing origin question and answer with potential extra contextual information. Contextual information includes the caption of the original images and OCR tokens for the scene-related question.
The prompt used for answer paraphrasing can be found in Appendix. Figure~\ref{fig:data_process} illustrates the data modifications we performed on our dataset.
\begin{figure}[t!]
    \centering
    \includegraphics[width=0.9\linewidth]{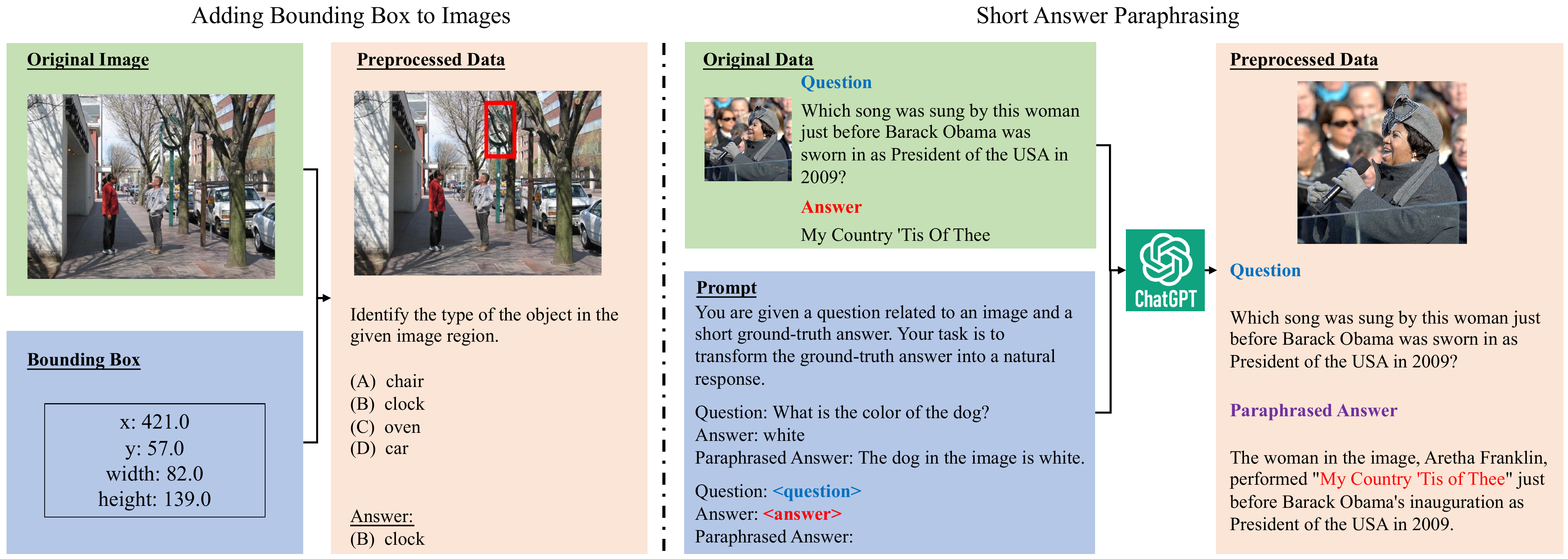}
    \caption{(Left) On region-based tasks, bounding boxes are added to original images to inform the model of the area in interest. (Right) Short answer paraphrasing to improve the response quality. 
    % \qi{need to explain a bit about why we should add the bounding box. Also, mention that this step is optional depending on the dataset.}
    }
    \label{fig:data_process}
\end{figure}

\noindent\textbf{Stage III: Quality Check}
% At the final stage, we assign a different annotator to each task for checking $10$ examples at each split.
% In this stage, we identified slight format mismatches between tasks, and resolved them by unifying the task formats. 
% We also noticed that a few answers (less than $3$ \% of examined instances) were not  paraphrased successfully by the ChatGPT, e.g., the response stated that ChatGPT could not paraphrase the original answer due to the lack of image information. 
% We use simple heuristics to filter these paraphrased answers and adopt a simple template to convert the original answer to a sentence. We notice almost no influence of this small portion of failed paraphrased answers.
% Finally,
% the task dataset is labeled as passed after the annotator can successfully load the dataset and re-examine the correctness of the instruction, inputs, and outputs of each instance. 
% \qi{typos here}
In this stage, we assign a different annotator to each task to review 10 examples from each split. During this stage, we identify minor format inconsistencies between tasks and address them by standardizing the task formats. We also observe that a few answers (less than 3\% of examined instances) were not effectively paraphrased by ChatGPT
% , such as responses stating that ChatGPT could not paraphrase the original answer 
due to insufficient image information. We employ simple heuristics to filter these paraphrased answers and use a basic template to convert the original answer into a sentence. We find that this small portion of unsuccessful paraphrased answers has negligible impact. Finally, the task dataset is deemed complete once the annotator can successfully load it and re-examine the accuracy of the instructions, inputs, and outputs for each instance examined.

\noindent\textbf{Stage IV: Key Datasets Translation} 
To boost the language diversity and support the evaluation across different languages, we select a subset of datasets~(OK-VQA, ImageNet, Winoground, VQAv2, VIST, MSRVTT and MSRVTT-QA) that covers different tasks and translate their evaluation data into $100$ languages following FLORES-101~\citep{DBLP:journals/tacl/GoyalGCCWJKRGF22}. 
We translate $500$ samples for each split of each task in our first version. 
More multilingual samples will be supported in the future.
We adopt the distillation version NLLB-1.3B~\citep{costa2022nllb} for translation, one of the state-of-the-art open multilingual translation models. 
As there are no native speakers for different languages, we adopt an automatic filtering mechanism to ensure the translation quality, where languages with translation BLEU scores from English larger than $20$ based on FLORES-101 results are kept. 
After this step, only $80$ languages are kept (see Appendix for detailed language names).
% Due to the extensive number of languages and computational resource constraints, only the written instructions have been translated at the time of submission. 
% We are committed to updating the dataset with complete translations.
% Due to the large number of languages and the computational resource limit, we only finished the translation of the written instructions upon the submission time. 
% We will update the dataset once the translation is finished.

\subsection{Dataset Format}
\label{subsec:data_format}

% \begin{table}[!ht]
% \small 
% \caption{A ViQuAE instance represented in the unified data instance schema used in our dataset.}
% \label{dataset_format}
% \begin{tcolorbox}
% \textcolor[rgb]{0,0.5,0.0}{\# List[String]:  the base64 string representation of a profile photo of F. Scott
% Fitzgerald}

% Images: \textcolor[rgb]{0,0,0.9}{["iVBORw0KGg...5ErkJggg=="]} 

% \textcolor[rgb]{0,0.5,0.0}{\# String: task instruction}

% Instruction: \textcolor[rgb]{0,0,0.9}{"Analyze the image and provide an appropriate response to the question."} 

% \textcolor[rgb]{0,0.5,0.0}{\# String: task-specific input, e.g., a question related to the image.}

% Input: \textcolor[rgb]{0.0,0,0.9}{"On which book by this man, Baz luhrmann’s planned a film?"} 

% \textcolor[rgb]{0,0.5,0.0}{\# String: task output, e.g., the correct answer for the question.}

% Output: \textcolor[rgb]{0,0,0.9}{"Baz Luhrmann has planned a film adaptation of the book The Great Gatsby."} 

% \textcolor[rgb]{0,0.5,0.0}{\# Dict: meta information dict contains original data.}  

\begin{figure}[ht!]
    \centering
\includegraphics[width=0.6\linewidth]{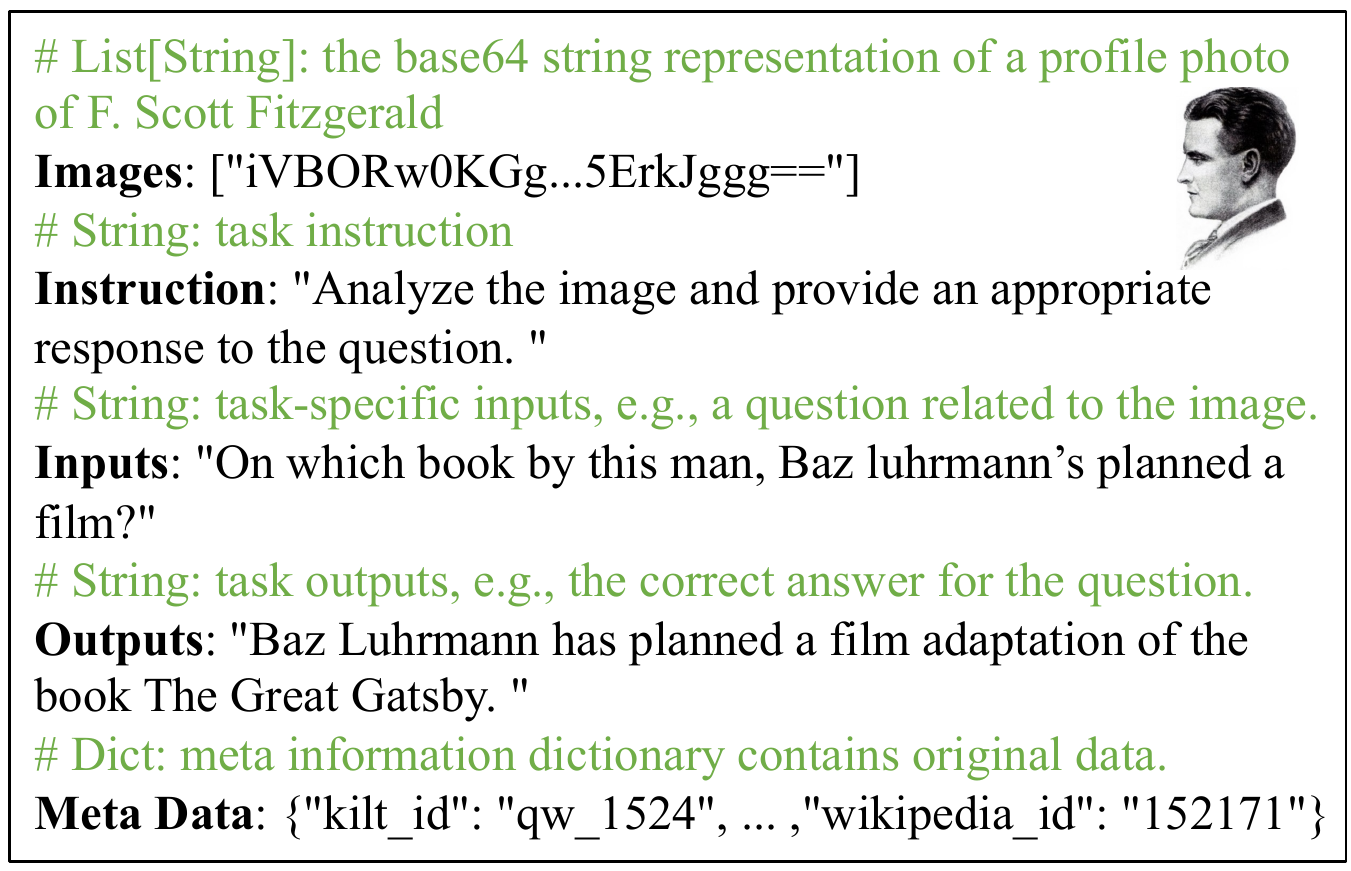}
    \caption{A ViQuAE instance represented in the unified data instance schema used in our dataset.}
    \label{fig:dataset_format}
\end{figure}

% Meta Data: \textcolor[rgb]{0,0,0.9}{\{"kilt\_id": "qw\_1524", ... ,"wikipedia\_id": "152171"\}} 
% \end{tcolorbox}
% \end{table}
The instance in our dataset consists of five fields: 
(1) \textbf{Images}: we represent the images with the potentially added bounding box by a base64 string. 
(2) \textbf{Instruction}: we randomly select an instruction from  the task instruction pool for each instance.
(3) \textbf{Inputs}: we allocate this field for providing task-specific inputs to the model, e.g., the question in the VQA tasks.
For tasks such as captioning, there is no extra input so the corresponding field is left as an empty string.
(4) \textbf{Outputs}: the required output to the specific tasks, such as the description of the image for captioning tasks and the answer to the image-related question.
(5) \textbf{Meta Data}: we provide this field to preserve important information such as image id for referencing the original dataset. 
Figure~\ref{fig:dataset_format} illustrates an instance in the unified format.
With the clear distinction of these fields, the user of our benchmark can flexibly construct the training instances needed and evaluate the models conveniently. 
Table~\ref{tab:task_desc_stat} gives the statistics aggregated by tasks, and we refer readers to Appendix for detailed statistics and the license of each dataset.

% \paragraph{Instructions} We evaluate the diversity of our instructions for different tasks and list the detailed statistics in Table~\ref{tab:instruct_stat}

\begin{table}[th!]
  \caption{M$^3$IT task descriptions and statistics, encompassing image captioning (CAP), classification (CLS), visual question answering (VQA), knowledgeable visual question answering (KVQA), reasoning (REA), generation (GEN), Chinese vision-language, and video-language tasks. We aggregate instance counts for training, validation, and test sets across all tasks, totaling 2,429,264 instances.}
  \label{tab:task_desc_stat}
  \centering
  \resizebox{0.9\linewidth}{!}{
  \begin{tabular}{clrrr}
    \toprule
    \multirow{2}{*}{Task} & \multirow{2}{*}{Description} & \multicolumn{3}{c}{Total \#samples} \\
    & & Train & Val & Test \\
    \midrule
    CAP & Given an image, write a description for the image. & 679,087 & 41,462 & 27,499 \\
    CLS & Given an image, classify the image into pre-defined categories. & 238,303 & 100,069 & 21,206 \\
    VQA  & Given an image, answer a question relevant to the image. & 177,633 & 46,314 & 10,828 \\
    KVQA & Given an image, answer the question requires outside knowledge. & 39,981 & 11,682 & 5,477 \\
    REA & Given an image, conduct reasoning over the images. & 99,372 & 11,500 & 10,000 \\
    GEN & Given an image, make compositions with certain requirements. & 145,000 & 11,315 & 17,350 \\
    \midrule
    Chinese & CAP, CLS, VQA, and GEN tasks in Chinese. & 192,076 & 77,306 & 4,100 \\
    Video & CAP, CLS, and VQA tasks on video-language datasets. & 20,868 & 7,542 & 9,294 \\
    \midrule
    % Multi-lingual we translate whole Winoground 800 * 40 
    % other set restricted to 500
    Multi-lingual & Translated tasks in $80$ languages &  0 & 240,000 & 184,000\\ 
    % \midrule 
    % Total & \multicolumn{4}{c}{ 2,005,264}
    % \\
    % w/ Chinese & Video: train 1,592,320; val 307,190; test 105,754
    % + multi-lingual = 
    % w/o Chinese & Video: train 1,379,376; val 222,342; test 92,360
    
    \bottomrule
  \end{tabular}
  \label{tab:dataset_statistics}
  }
\end{table}

\section{Experiments}
In this section, we build a VLM to validate the effectiveness of the proposed M$^3$IT dataset for multi-modal agents. We first introduce the experimental setups~(\S~\ref{subsec:exp_setup}), then report and discuss the results~(\S~\ref{subsec:main_ret}). Lastly, we analyze the influence of task number and instruction diversity, and provide a qualitative result~~(\S~\ref{subsec:analysis_ret}).
% \qi{briefly introduce each section.}

\subsection{Experimental Settings}
\label{subsec:exp_setup}

\noindent\textbf{Implementation Details} Inspired by the recent success of BLIP~\cite {li2023blip2}, we adopt the vision encoder and the Q-former architecture in the BLIP2-OPT-2.7B~\citep{li2023blip2} model to extract relevant visual features from images. For the large language models, we utilize Ziya-13B~\citep{fengshenbang} derived from LLaMA~\citep{touvron2023llama} with bilingual~(English and Chinese) ability. 
We employ a two-staged training.
\textbf{Stage I Visual-Text Alignment:} To align the visual and textual feature space, we utilize the instructions in the coco captioning and perform an initial alignment training on LAION 400M~\citep{laion400m}.
We train the Q-former and the language projection, resulting in a total $130$M parameters to optimize with AdamW~\citep{loshchilov2018decoupled}.
The batch size is set to $256$ to maximize the utilization of GPU and the model is trained with $300$k steps. The learning rate linearly increases to a peak value of 5$e$-5 in the first $2000$ steps and follows a cosine decay scheduler. The weight decay is set to $0.05$.
\textbf{Stage II Multi-modal Instruction Tuning:}
We further perform a multi-modal instruction tuning in our benchmark to activate the great potential of LLMs.
We train the model after alignment training for $3$ epochs and with a 
lower learning rate of 1$e$-5 and a warmup stage of $1000$ steps. 
Inspired by LoRa tuning~\citep{hu2021lora}, the weights for mapping query and value vectors in the attention layer of LLMs are learnable in this stage to better adapt to the instruction tuning dataset.
Other training parameters are consistent with Stage I. All experiments are conducted with 8 NVIDIA 80GB A100 GPUs. It took about 10 days for Stage I and Stage II can be finished in a day.

\noindent\textbf{Evaluation Setup} 
% Our dataset is built for investigating the cross-task generalization performance after instruction tuning for vision-language models.
To examine the generalization of instruction tuning,  some tasks are held-out for evaluation (see Figure~\ref{fig:dataset_tasks} for held-in/out tasks). 
% Specifically, we evaluate three challenging setups:
We are interested in the following research questions:
(RQ1) Can multi-modal instruction tuning elicit world knowledge from LLMs? 
(RQ2) Can English-only instruction tuning generalize to other languages such as Chinese?
and (RQ3) Can image-only multi-modal instruction tuning generalize to video-language tasks?
For RQ1, we evaluate our models on three KVQA tasks in our datasets, i.e., OK-VQA~\citep{marino2019okvqa}, A-OKVQA~\citep{schwenk2022aokvqa} and ViQuAE. 
For RQ2 and RQ3, we perform zero-shot transfer evaluation on Chinese vision-language and video-language datasets, respectively.
We use greedy decoding in inference if not otherwise specified.

% \qi{briefly add some motivations of using this. Say that we mainly focus on evluating the conversational abilities of the VLM model.}
\noindent\textbf{Metrics} 
We adopt ROUGE-L~\citep{lin-2004-rouge} as an automatic metric to assess the consistency between predictions and ground-truth answers, focusing on evaluating the model's conversational abilities.
As the automatic metric may not fully capture the nuances of conversational quality, we further introduce GPT-4 as a proxy of human evaluators~(\S~\ref{subsec:gpt4_eval}). 

\noindent\textbf{Baselines} 
We compare our models to recently proposed powerful multi-modal agents, including (1) BLIP-2-Flan-T5-XXL~\citep{li2023blip2} where an instruction-tuned Flan-T5~\citep{weifinetuned} is connected with a powerful vision encoder to perform a series of multi-modal tasks; (2) MiniGPT-4 which aligns a CLIP visual encoder with a frozen Vicuna~\citep{vicuna2023} with artificially collected dialog dataset; and (3) InstructBLIP, a recently proposed instruction tuning enhanced multi-modal agents with Vicuna-13B with converted multi-model datasets and the LLaVA~\citep{liu2023llava} dataset generated by GPT-4.

% \textbf{From In-Domain Distribution to Out-of-Domain Distribution}

% \textbf{From English to Other Languages}

% \textbf{From Images to Video}

% Across-Language 
% Cross-Modality

% For held-out evaluation, our aim is to understand how instruction tuning improves the model’s zeroshot generalization performance on unseen data. In this paper, we define two types of held-out data:
% 1) datasets not exposed to the model during training, but whose tasks are present within the held-in
% cluster; 2) datasets and their associated tasks that remain entirely unseen during the training process.
% Addressing the first type of held-out evaluation is nontrivial due to image distribution shift between
% held-in and held-out datasets. As for the second type, we hold out several tasks completely, including
% visual reasoning, video question answering, visual conversational QA, and image classification

\subsection{Main Results}
\label{subsec:main_ret}
% \qi{need to include metrics in the talbes or figures. A bit confusing about which metric is used.}
% \input{tables/ood_ret}
% \begin{table}[t!]
%   \centering
%   \caption{ROUGE-L evaluation results of Chinese vision-language tasks.}
%     \small 
%   \begin{adjustbox}{max width=\textwidth}
%   \begin{tabular}{lccc}
%     \toprule
%     Model & Flickr-CN& FM-IQA & Chinese-FoodNet \\ 
%     \midrule
%     MiniGPT4 & 9.6 & 20.1&5.0 \\
%     InstructBLIP  & 5.2 & 2.3& 1.0 \\
%     Ours & \textbf{20.5}& \textbf{33.3}& \textbf{49.8} \\
%     \bottomrule
%   \end{tabular}
%   \end{adjustbox}
%   \label{tab:zs_chinese}

% \end{table}

\begin{table}[t!]
  \centering
  \begin{minipage}[b]{0.49\textwidth}
  \centering
  \caption{
  ROUGE-L evaluation results of KVQA tasks.  Our Ying-VLM outperforms all the baselines consistently.
  % Experiment result on knowledgeable visual question answering benchmarks. Our Ying-VLM outperforms all the baseline models with consistently better ROUGE-L scores.
  }
  \begin{adjustbox}{max width=\textwidth}
  \begin{tabular}{lccc}
    \toprule
    Model &  OK-VQA & A-OKVQA & ViQuAE  \\ 
    \midrule
    BLIP2-Flan-T5-XXL & 9.1  & 15.6  & 9.7  \\ 
    MiniGPT4 & 23.3   & 21.8 & 24.4 \\
    InstructBLIP  & 7.1   & 5.9  & 7.3  \\
    Ying-VLM (Ours) & \textbf{27.5}  & \textbf{24.5}  & \textbf{29.6}  \\
    \bottomrule
  \end{tabular}
  \end{adjustbox}
  \label{tab:kvqa_ret}

  \end{minipage}
  \hfill
  \begin{minipage}[b]{0.45\textwidth}

      \centering
  \caption{Zero-shot transfer to Chinese vision-language tasks. Our model generalizes well on unseen Chinese captioning, VQA and classification tasks, with the highest ROUGE-L.}
  \begin{adjustbox}{max width=\textwidth}
  \begin{tabular}{lccc}
    \toprule
    Model & Flickr-8k-CN& FM-IQA & Chinese-FoodNet \\ 
    \midrule
    MiniGPT4 & 9.6 & 20.1&5.0 \\
    InstructBLIP  & 5.2 & 2.3& 1.0 \\
    Ying-VLM (Ours) & \textbf{20.5}& \textbf{33.3}& \textbf{49.8} \\
    \bottomrule
  \end{tabular}
  \end{adjustbox}
  \label{tab:zs_chinese}
  \end{minipage}
\end{table}

\noindent\textbf{RQ1: Knowledgeable Visual Question Answer Evaluation}
% We are interested in whether the instruction tuning can activate the knowledge in the large language models by connecting the visual content with the textual spaces of language models.
% To examine this, we propose to evaluate models on two knowledgeable datasets OK-VQA and A-OKVQA. 
% We also perform a zero-shot evaluation, i.e., the training dataset is not used in our Stage II instruction tuning, on an entity-centric knowledgeable question-answering dataset, ViQuAE.
The results on the KVQA benchmarks are shown in Table~\ref{tab:kvqa_ret}.
In comparison to the strongest baseline, our model achieves an improvement of 3.2 and 2.7 ROUGE-L points for OK-VQA and A-OKVQA, respectively.
Additionally, Ying-VLM delivers the best performance on the held-out ViQuAE dataset. These findings indicate that instruction tuning on M$^3$IT effectively harnesses knowledge from LLMs and elevates response quality.

% While InstructBLIP also incorporates OK-VQA and A-OKVQA tasks into the training dataset, the score is sacrificed due to the fact that InstructBLIP tends to produce short answers, as revealed in our later quantitative results~(\S\ref{subsec:case_study}). 
% \qi{consider removing this argument about instructblip. there might be other reasons that we perform better like more datasets, better response quality etc. This argument reduces the significance of the result.} 

\noindent\textbf{RQ2: Zero-Shot Transfer to Chinese Vision-Language Tasks}
We assess models on three unseen Chinese vision-language tasks to investigate the cross-language generalization capabilities of instruction tuning. 
BLIP-2 is not considered, as Flan-T5 does not support Chinese.\footnote{For all models, we introduce a prompt to promote Chinese outputs. See Appendix D for details.} 
As illustrated in Table~\ref{tab:zs_chinese}, our model outperforms MiniGPT4 and InstructBLIP on all evaluated tasks, demonstrating notable improvements. These findings indicate that instruction tuning with English datasets can effectively generalize to different languages, showcasing the promising potential that can be further explored.

\begin{table}[t!]
  \centering
  \caption{Zero-shot transfer to video-language tasks. We report ROUGE-L score for all tasks.}
  \small 
  \begin{adjustbox}{max width=\textwidth}
  \begin{tabular}{lccccc}
    \toprule
    \multirow{2}{4em}{Model} & Video Captioning & \multicolumn{4}{c}{Video Question Answer} \\ \cmidrule(lr){3-6} %\cmidrule(lr){2} \cmidrule(lr){3-6}
    & MSRVTT & iVQA & ActivityNet-QA & MSRVTT-QA & MSVD-QA \\
    \midrule
    BLIP-2-Flan-T5-XXL & 8.8 & 11.1 & 8.9 & 10.3 & 13.2 \\
    % MiniGPT4 & 11.8 & 42.3 & 41.8 & 28.8 & 40.3 \\
    InstructBLIP & \textbf{14.3} & 6.3 & 9.3 & 4.0 & 7.0 \\
    Ying-VLM (Ours) & 14.2 & \textbf{23.5} & \textbf{21.9} & \textbf{18.3} & \textbf{21.4} \\
    \bottomrule
  \end{tabular}
  \end{adjustbox}
  \label{tab:zs_video}
\end{table}
\noindent\textbf{RQ3: Zero-Shot Transfer to Video-Language Tasks} To evaluate performance on video-language tasks, we uniformly sample $8$ frames from each video. A comparison with MiniGPT4 is excluded, as it does not support video inputs. Following the approach of InstructBLIP~\citep{dai2023instructblip}, we concatenate the visual embedding extracted from the Q-former of each frame as a prefix embedding to the language model. As demonstrated in Table~\ref{tab:zs_video}, our model excels in these challenging settings, significantly surpassing the BLIP-series baselines. It is worth noting that the training dataset does not include any visual inputs such as videos, implying that our instruction tuning effectively aids the model in generalizing to video inputs with a temporal dimension.

% We further explore whether the learned models can generalize to the video. 
% To perform the evaluation on video-language tasks,
% we uniformly sample $8$ frames from each video. We do not compare it with MiniGPT4 as it does not support video inputs.
% Following InstructBLIP~\citep{dai2023instructblip}, the visual embedding extracted from the Q-former of each image is concatenated as a prefix embedding to the language model. 
% As shown Table~\ref{tab:zs_video}, our models can transfer to these challenging settings, significantly outperforming BLIP-series baselines.
% Note that the training dataset does not contain any visual inputs like videos, suggesting that our instruction tuning helps the model generalize to video inputs with the temporal dimension.
% We also notice that InstructBLIP achieves decent performance on the image captioning task, while performing relatively poorly on the more challenging video question answering.

\begin{figure}[ht!]
    \centering
\includegraphics[width=0.6\linewidth]{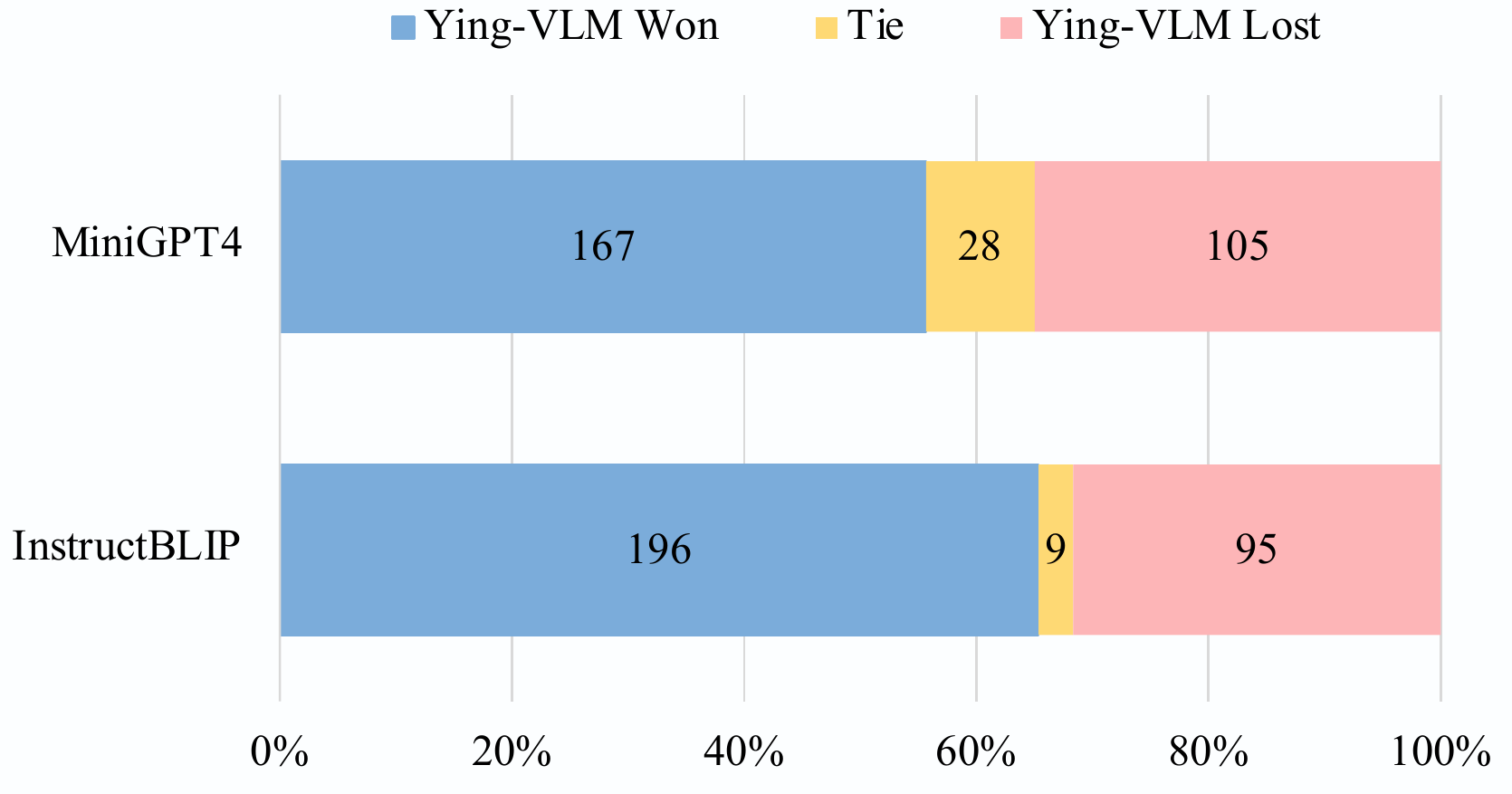}
    \caption{Evaluation results using GPT-4 as an evaluator. Our model outperforms MiniGPT-4 and InstructBLIP with a winning rate at 55.6\% and 65.5\%, respectively.}
    \label{fig:gpt-4-eval}
\end{figure}
\noindent\textbf{GPT-4 Evaluation Results}
\label{subsec:gpt4_eval}
To further validate the quality of the generated response, we propose to utilize the powerful GPT-4 model as a proxy of human evaluators~\citep{it-with-gpt4,gilardi2023chatgpt}.
Specifically, following Vicuna~\cite{vicuna2023}, we use GPT-4 to rate the performance of different models against our Ying-VLM.
Considering the API cost of GPT-4, $300$ examples are randomly sampled from OK-VQA, A-OKVQA and ViQuAE datasets as a subset for evaluation.
For each sample, we construct a prompt consisting of the original question, its corresponding reference answer, the response generated by our Ying-VLM, and a baseline system output. 
GPT-4 is queried with the prompt to rate both responses on a scale of ten based on the given question and its reference answer.
The ratings are primarily based on the accuracy, relevance, and naturalness of the response to meet the requirements when humans are interacting with multi-modal agents (see Appendix for the detailed evaluation template).
We employ the strategy proposed by \citet{wang2023large} to mitigate potential evaluation biases regarding the response order.\footnote{\url{https://github.com/i-Eval/FairEval}}
Figure~\ref{fig:gpt-4-eval} shows that our Ying-VLM outperforms all baseline models in most samples. 
Notably, Ying-VLM beat the strongest baseline MiniGPT4 on $167$ over $300$ tested samples.
Consistent with the previous ROUGE-L evaluation, this result indicates that the model fine-tuned on our instruction dataset can produce more accurate and engaging responses on the challenging KVQA tasks.

\subsection{Analysis}
\label{subsec:analysis_ret}
We investigate the effect of task number and instruction diversity on the performance of learned models, providing insights for future studies to utilize our benchmark better.

\begin{figure}[htbp]
\centering
\begin{minipage}[b]{0.48\textwidth}
    \centering
    \includegraphics[width=0.9\textwidth]{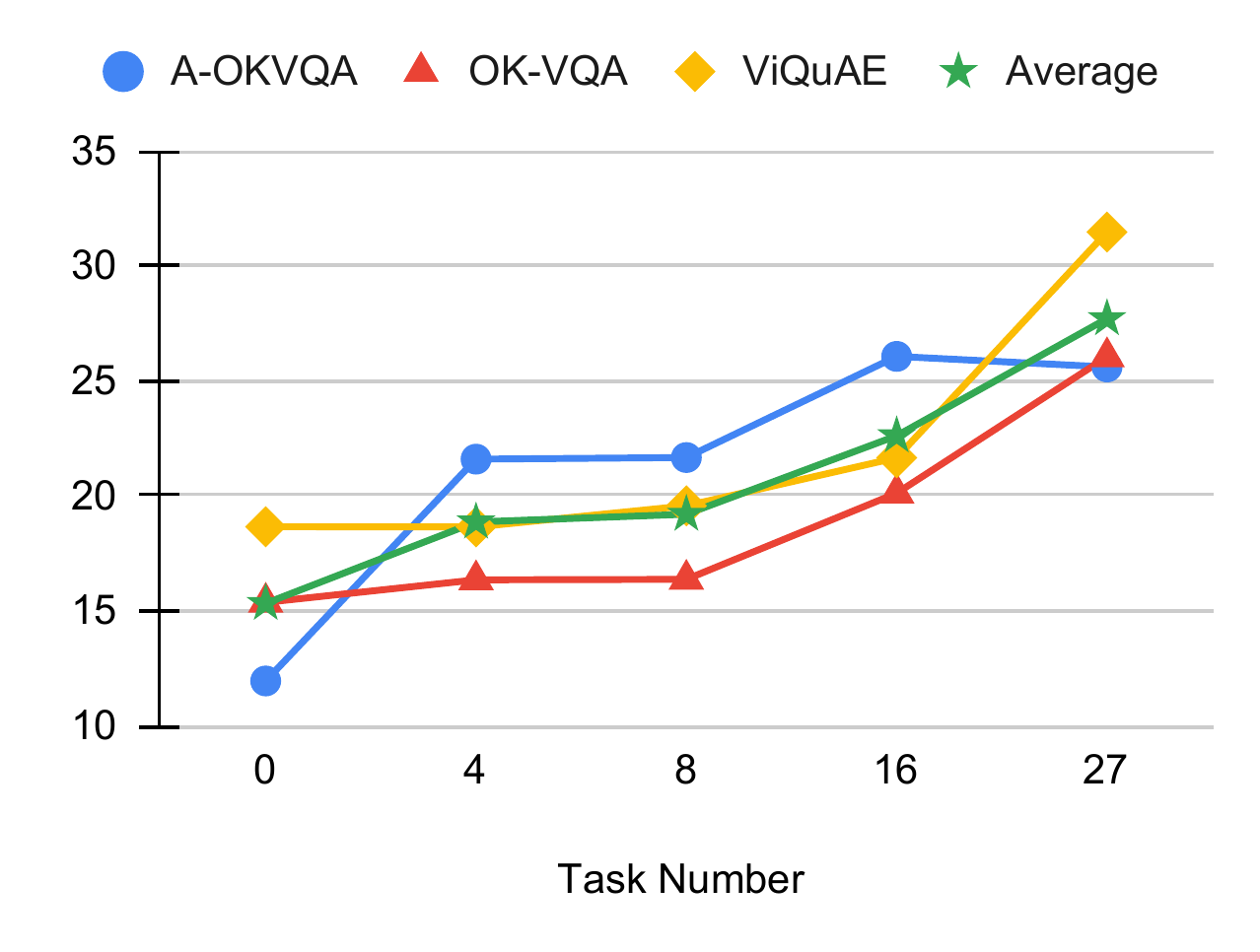}
    \caption{Performance increases with more instruction tuning datasets.}
    \label{fig:vary_task_num}
\end{minipage}
\hfill
\begin{minipage}[b]{0.48\textwidth}
    \centering
    \includegraphics[width=0.9\textwidth]{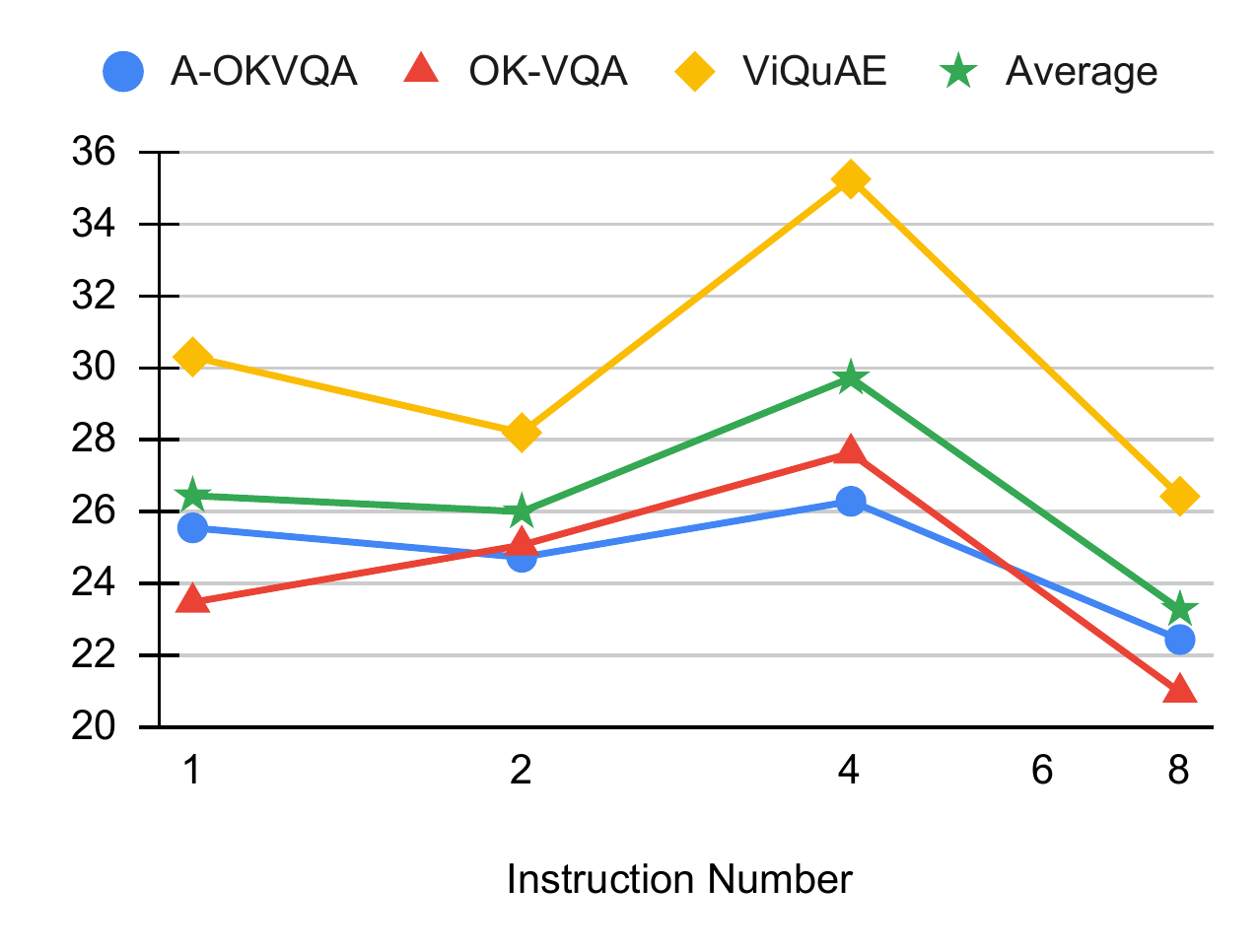}
    \caption{Performance changes with the varied number of instructions used for training.}
    \label{fig:vary_instruct_num}
\end{minipage}
\end{figure}

\noindent\textbf{Effect of Task Number}
We investigate the influence of task numbers by randomly shuffling our tasks and then selecting a subset to train the model during the instruction tuning stage.
Due to the computational resource limitation, we set up a maximum of $5$k examples for each task and train all the models for $5$k steps with a batch size of $64$.
We select $0$, $4$, $8$, $16$ and all $27$ tasks for training, and report the individual ROUGE-L score and the average score.
As illustrated in Figure~\ref{fig:vary_task_num}, increasing the number of tasks greatly improves the results of the generalization performance.
% a dataset not appearing in the training datasets. 
Besides, the performance gain is not diminished as the task number increases. This is promising as it indicates that we can continually improve performance by introducing more tasks into the training.
It would be interesting to investigate the influence of different task clusters, which we leave for future studies.

\noindent\textbf{Effect Instruction Diversity} 
To investigate the influence of instruction diversity, we limit the number of instructions used in each dataset to $1$, $2$, $4$, and $8$, resulting in varying levels of diversity for each task. The other training parameters are consistent with those used in previous experiments on task number investigation. Figure~\ref{fig:vary_instruct_num} shows that the performance varies with the level of diversity. Specifically, our results suggest that using four instructions per task is sufficient for achieving decent performance. 
% The results tested on the unseen task instructions also exhibit a similar trend~(Figure~1 in Appendix).
We leave a more in-depth analysis of the instruction diversity for future work.
% and how to develop a better framework to identify the most important instructions 

% Interestingly, the performance reaches a peak value when only using $4$ instructions per task.
% The ViQuAE task is smilar with OK and A-OKVQA ?

% We evaluate the generalization on two tasks
% (1) generalization to unseen task instruction, where we evaluate the performance on the instances with instructions that do not appear in training.

% (2) generalization to unseen tasks.

% \paragraph{Generalize to Unseen Instructions}

\noindent\textbf{Qualitative Results}
\label{subsec:case_study}
% picking some examples 
\begin{figure}
    \centering
\includegraphics[width=0.9\linewidth]{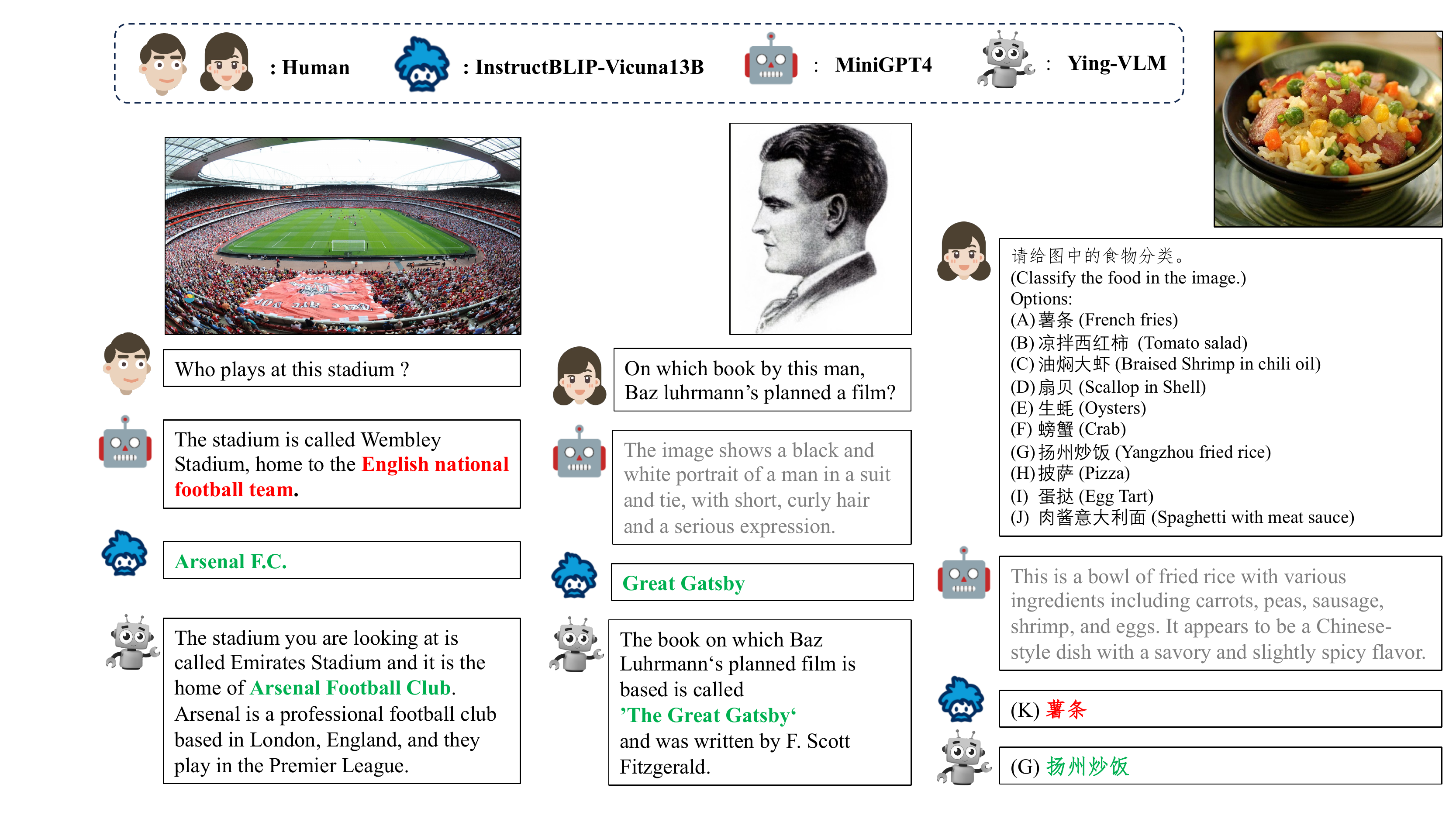}
    \caption{Case study of the model outputs. Correct answers are bolded with green, wrong answers are in red and irrelevant answers are in grey. The model trained with our datasets can provide natural and informative responses to entity-centric questions, and generalize to the food classification task in Chinese (English translation for visualization only). }
    \label{fig:case_study}
\end{figure}
We conduct a case study to provide a more straightforward understanding of instruction-tuned models. The cases are chosen from the held-out ViQuAE and ChineseFoodNet datasets. 
As shown in Figure~\ref{fig:case_study}, our model generates accurate responses to all questions. In contrast, MiniGPT4 produces an incorrect answer for the stadium question on the left and fails to follow instructions in the subsequent cases, providing generic image descriptions instead. Additionally, compared to InstructBLIP, which provides concise but less engaging answers for the two questions requiring external knowledge, our model responds more naturally and engagingly, underlining the value of our dataset. 
Our model also successfully generalizes to Chinese inputs, accurately classifying the food image based on the instruction. These cases emphasize the importance of instruction tuning and demonstrate that our dataset can effectively enhance the capabilities of VLMs.

\section{Conclusion}
In this paper, we present M$^3$IT, a multi-modal multilingual instruction tuning dataset for aiding the development of multi-modal large language models.
The dataset comprises 2.4 million carefully curated instances and 400 manually written task instructions across 40 tasks.
We build Ying-VLM to validate the effectiveness of our dataset.
Quantitative and qualitative results demonstrate that the models trained with our datasets successfully follow human instructions, provide more engaging responses, and achieve strong generalization performance on unseen video and Chinese tasks. 
Further analysis shows that the increased task number can continually boost performance, and instruction diversity can influence results.
We hope our proposed benchmark, trained models, and experimental findings can facilitate future studies toward building powerful multi-modal intelligent agents.

\appendix

\section{Dataset Statistics}
\label{apx:dataset_statistics}

% A detailed table showing each dataset in Appendix
\begin{table}[ht!]
  \caption{Detailed task descriptions and statistics of our instruction tuning tasks, including all datasets in all types of tasks. The column ``Used'' indicates whether we use this dataset in the instruction tuning stage.}
  \label{tab:dataset_statistics_all}
  \centering
  \small 
  \begin{adjustbox}{max width=\textwidth}
  \begin{tabular}{c|c|c|ccc|c}
    \toprule
    \multirow{2}{*}{Task} & \multirow{2}{*}{Dataset} & \multirow{2}{*}{Used} & \multicolumn{3}{c|}{\#samples} & \multirow{2}{*}{License}\\
    & & & Train & Val & Test & \\
    \midrule
    \multirow{3}{*}{Captioning} & MS COCO~\citep{lin2014mscoco} & Yes & 566,747 & 25,010 & 25,010  & Custom\\
    & TextCaps~\citep{sidorov2020textcaps} & Yes & 97,765 & 13,965 & 0 & Unknown \\
    & Image-Paragraph-Captioning~\citep{krause2016image_para_cap} & Yes & 14,575 & 2,487 & 2,489 & Custom\\
    \midrule
    \multirow{7}{*}{Classification} & COCO-GOI~\citep{lin2014mscoco} & Yes & 30,000 & 2,000 & 0 & Custom\\
    & COCO-Text~\citep{veit2016coco_text} & Yes & 118,312 & 27,550 & 0 & Custom \\
    & ImageNet~\citep{russakovsky2015imagenet} & Yes & 30,000 & 50,000 & 0 & Non-commercial \\
    & COCO-ITM~\citep{lin2014mscoco} & Yes & 30,000 & 5,000 & 5,000 & Custom \\
    & e-SNLI-VE~\citep{esnlive} & Yes & 20,000 & 14,339 & 14,740 & Unknown\\
    & Mocheg~\citep{yao2022mocheg} & Yes & 4,991 & 180 & 466 & CC BY 4.0\\
    & IQA~\citep{duanmu2021iqa} & Yes & 5,000 & 1,000 & 1,000 & Custom\\
    \midrule
    \multirow{7}{*}{VQA} & VQA v2~\citep{balanced_vqa_v2} & Yes & 30,000 & 30,000 & 0 & CC-BY 4.0 \\
    & Shapes VQA~\citep{andreas2016shapes} & Yes & 13,568 & 1,024 & 1,024 & Unknown \\
    & DocVQA~\citep{mathew2021docvqa} & Yes & 39,463 & 5,349 & 0 & Unknown\\
    & OCR-VQA~\citep{mishra2019ocr_vqa} & Yes & 11,414 & 4,940 & 0 & Unknown \\
    & ST-VQA~\citep{biten2019st_vqa} & Yes & 26,074 & 0 & 4,070 & Unknown\\
    & Text-VQA~\citep{singh2019text_vqa} & Yes & 27,113 & 0 & 5,734 &  CC BY 4.0\\
    & GQA~\citep{hudson2019gqa} & Yes & 30,001 & 5,001 & 0 & CC BY 4.0 \\
    \midrule
    \multirow{4}{*}{KVQA} & OK-VQA~\citep{marino2019okvqa} & Yes & 9,009 & 5,046 & 0 & Unknown\\
    & A-OK-VQA~\citep{schwenk2022aokvqa} & Yes & 17,056 & 1,145 & 0 & Unknown\\
    & ScienceQA~\citep{lu2022scienceqa} & Yes & 12,726 & 4,241 & 4,241 & CC BY-NC-SA\\
    & ViQuAE~\citep{lerner2022viquae} & No & 1,190 & 1,250 & 1,236 & CC By 4.0\\
    \midrule
    \multirow{5}{*}{Reasoning} & CLEVR~\citep{johnson2017clevr} & Yes & 30,000 & 2,000 & 0 & CC BY 4.0\\
    & NLVR~\citep{Suhr2017NLVR} & Yes & 29,372 & 2,000 & 0 & Unknown\\
    & VCR~\citep{zellers2019vcr} & Yes & 25,000 & 5,000 & 5,000 & Custom \\
    & VisualMRC~\citep{tanaka2021visualmrc} & Yes & 15,000 & 2,500 & 5,000 & Unknown \\
    & Winoground~\citep{thrush2022winoground} & No & 0 & 0 & 800 & Unknown \\
    \midrule
    \multirow{3}{*}{Generation} & Visual Storytelling~\citep{huang2016vist} & Yes & 5,000 & 4,315 & 4,350 & Unknown \\
    & Visual Dialog~\citep{das2017visual_dialog} & Yes & 50,000 & 1,000 & 1,000 & CC By 4.0 \\
    & Multi30k~\citep{elliott2016multi30k} & Yes & 90,000 & 6,000 & 12,000 & Non-commercial\\
    \midrule
    \multirow{5}{*}{Chinese} & FM-IQA~\citep{fm-iqa} & No & 164,735 & 75,206 & 0 & Unknown \\
    & COCO-Caption CN~\citep{li2019cococn} & No & 18,341 & 1,000 & 1,000 & Non-commercial \\
    & Flickr-8k-Caption CN~\citep{flickr8kcn} & No & 6,000 & 1,000 & 1,000 & CC By 3.0 \\
    & Chinese Food Classification~\citep{chen2017chinesefoodnet} & No & 0 & 0 & 1,100 & Unknown \\
    & Multimodal Chat~\citep{zheng2021mmchat} & No & 3,000 & 1,000 & 1,000 & Unknown \\
    \midrule
    \multirow{6}{*}{Video} & Action-Classification~\citep{goyal2017something_something} & No & 2,000 & 2,000 & 2,000 & Custom \\
    & iVQA~\citep{yang2021iVQA} & No & 5,994 & 2,000 & 2,000 & Unknown \\
    & MSVD QA~\citep{xu2017msrvtt_qa} & No & 1,161 & 245 & 504 & Unknown \\
    & ActivityNet QA~\citep{yu2019activitynet} & No & 3,200 & 1,800 & 800 & Unknown \\
    & MSRVTT QA~\citep{xu2017msrvtt_qa} & No & 6,513 & 497 & 2,990 & Unknown \\
    & MSRVTT Captioning~\citep{xu2016msrvtt} & No & 2,000 & 1,000 & 1,000 & Unknown \\
    \bottomrule
  \end{tabular}
  \end{adjustbox}
\end{table}
Table~\ref{tab:dataset_statistics_all} lists the detailed statistics in our benchmark.
We collect the dataset license from PaperWithCode.\footnote{\url{https://paperswithcode.com/}}
For datasets under Unknown and Custom licenses, we suggest the users check the project page or contact the dataset owner before usage.

\section{Template for Answer Paraphrase}
\label{apx:paraphrase_template}

We provide the paraphrase template in Table~\ref{tab:chatgpt_paraphrase} for querying the ChatGPT to re-write the original short answers, where \textcolor[rgb]{0,0.0,0.9}{\{Q\}} and \textcolor[rgb]{0,0.0,0.9}{\{A\}} is filled with the question and the answer need to be paraphrased, respectively.
We incorporate an example to better inform the model of the paraphrasing tasks.
For VQAv2 tasks, we add an extra \textcolor[rgb]{0,0.0,0.9}{\{Caption\}} field in the template filled with corresponding captions from the COCO dataset to provide extra context information to help to paraphrase.

\begin{table}[ht!]
\caption{Template used to query ChatGPT for answer paraphrasing.}

\begin{tcolorbox}

You are an AI visual assistant. Now you are given a question related to an image and a short ground-truth answer.  
Your task is to transform the ground-truth answer into a natural and convincing response. 
Make sure the response is accurate, highly relevant to the question, and consistent with the original answer. 

\ 

Question: 

Which NASA space probe was launched to this planet in 1989?

Answer:

Magellan

Transformed Answer:

NASA sent the Magellan spacecraft to Venus in 1989, which was the first planetary spacecraft launched from a space shuttle.

\

Question:

\textcolor[rgb]{0,0.0,0.9}{\{Q\}}

Answer:

\textcolor[rgb]{0,0,0.9}{\{A\}}

Transformed Answer:

\end{tcolorbox}

\label{tab:chatgpt_paraphrase}
\end{table}

\section{Dataset Translation}
\label{apx:data_translation}
\begin{table}[ht!] 
\centering 
\caption{List of Language Codes, Scripts, and Languages Names for translated datasets.}
\tiny 
\begin{tabular}{lcc} 
\toprule
\textbf{Language Code} & \textbf{Script} & \textbf{Language Name} \\
\midrule
af & afr\_Latn & Afrikaans \\
am & amh\_Ethi & Amharic \\
ar & arb\_Arab & Modern Standard Arabic \\
as & asm\_Beng & Assamese \\
ast & ast\_Latn & Asturian \\
be & bel\_Cyrl & Belarusian \\
bg & bul\_Cyrl & Bulgarian \\
bn & ben\_Beng & Bengali \\
bs & bos\_Latn & Bosnian \\
ca & cat\_Latn & Catalan \\
ceb & ceb\_Latn & Cebuano \\
cs & ces\_Latn & Czech \\
cy & cym\_Latn & Welsh \\
da & dan\_Latn & Danish \\
de & deu\_Latn & German \\
el & ell\_Grek & Greek \\
es & spa\_Latn & Spanish \\
et & est\_Latn & Estonian \\
fi & fin\_Latn & Finnish \\
fr & fra\_Latn & French \\
fuv & fuv\_Latn & Nigerian Fulfulde \\
gl & glg\_Latn & Galician \\
gu & guj\_Gujr & Gujarati \\
ha & hau\_Latn & Hausa \\
he & heb\_Hebr & Hebrew \\
hi & hin\_Deva & Hindi \\
hr & hrv\_Latn & Croatian \\
hu & hun\_Latn & Hungarian \\
hy & hye\_Armn & Armenian \\
id & ind\_Latn & Indonesian \\
ig & ibo\_Latn & Igbo \\
is & isl\_Latn & Icelandic \\
it & ita\_Latn & Italian \\
ja & jpn\_Jpan & Japanese \\
jv & jav\_Latn & Javanese \\
ka & kat\_Geor & Georgian \\
kk & kaz\_Cyrl & Kazakh \\
km & khm\_Khmr & Khmer \\
kn & kan\_Knda & Kannada \\
ko & kor\_Hang & Korean \\
ky & kir\_Cyrl & Kyrgyz \\
lb & ltz\_Latn & Luxembourgish \\
lg & lug\_Latn & Ganda \\
lij & lij\_Latn & Ligurian \\
li & lim\_Latn & Limburgish \\
ln & lin\_Latn & Lingala \\
lo & lao\_Laoo & Lao \\
lt & lit\_Latn & Lithuanian \\
lv & lvs\_Latn & Standard Latvian \\
mi & mri\_Latn & Maori \\
mk & mkd\_Cyrl & Macedonian \\
ml & mal\_Mlym & Malayalam \\
mr & mar\_Deva & Marathi \\
mt & mlt\_Latn & Maltese \\
my & mya\_Mymr & Burmese \\
nl & nld\_Latn & Dutch \\
ny & nya\_Latn & Nyanja \\
oc & oci\_Latn & Occitan \\
pa & pan\_Guru & Eastern Panjabi \\
pl & pol\_Latn & Polish \\
pt & por\_Latn & Portuguese \\
ro & ron\_Latn & Romanian \\
ru & rus\_Cyrl & Russian \\
sd & snd\_Arab & Sindhi \\
sk & slk\_Latn & Slovak \\
sn & sna\_Latn & Shona \\
so & som\_Latn & Somali \\
sr & srp\_Cyrl & Serbian \\
sv & swe\_Latn & Swedish \\
ta & tam\_Taml & Tamil \\
te & tel\_Telu & Telugu \\
tg & tgk\_Cyrl & Tajik \\
th & tha\_Thai & Thai \\
tl & tgl\_Latn & Tagalog \\
tr & tur\_Latn & Turkish \\
uk & ukr\_Cyrl & Ukrainian \\
ur & urd\_Arab & Urdu \\
vi & vie\_Latn & Vietnamese \\
wo & wol\_Latn & Wolof \\
zh & zho\_Hans & Chinese (Simplified) \\ 
\bottomrule
\end{tabular} 
\label{tab:lan_code} 
\end{table}
We translate all the task instructions and evaluation sets of ImageNet, Winoground, VQAv2, OK-VQA, VIST, MSRVTT and MSRVTT-QA into $80$ languages, as shown in Table~\ref{tab:lan_code}.
Due to the computational resource constraint, we translate the whole test of Winoground ( $800$ examples) and set a maximum instance number of $500$ for each split in other tasks.
% \section{Evaluation on Unseen Instructions}

% \begin{figure}[ht!]
%     \centering
%     \includegraphics[width=0.5\linewidth]{images/instruct_number_ood.pdf}
%     \caption{Performance evaluated on the unseen task instructions}
%     \label{fig:ood_vary_instruct_num}
% \end{figure}

% We perform an extra analysis of the instruction diversity by examining the performance on the unseen task instructions. Specifically, we tested the generalization on the task instructions not used for training, with the models trained on different numbers of instructions. As shown in Figure~\ref{fig:ood_vary_instruct_num}, the performance changes and exhibit a similar trend with the results in the main paper.
% This indicates that instruction diversity also has an influence on generalization performance.

\section{Prompt for Zero-Shot Chinese Vision-Language Tasks}

In our experiments, all Vision-Language models are fine-tuned exclusively using English data. In our preliminary study, we observe that these models tend to generate English responses, even when the input and instructions are written in Chinese. 
We introduce a simple Chinese dialogue context during the zero-shot Chinese Vision-Language Task evaluation for all models, as illustrated in Table~\ref{tab:chinese_prompt}, 
Interestingly, this minor adjustment can encourage models to produce reasonable Chinese output. We leave the analysis of instruction-tuned VLM models' multilingual capabilities for future research.

\begin{table}[ht!]
\caption{Prompt for promoting Chinese outputs.}

\begin{tcolorbox}
\begin{CJK*}{UTF8}{gbsn} % 进入 CJK 环境，gbsn 字体为宋体
<human>:
请根据我的指示，以及所给的图片，做出相应的回答。

<bot>:

好的。

<human>:

\textcolor[rgb]{0,0.0,0.9}{\{Instruction\}}

\textcolor[rgb]{0,0.0,0.9}{\{Input\}}

<bot>:

好的。

\end{CJK*}

\end{tcolorbox}

\label{tab:chinese_prompt}
\end{table}

\section{Template for GPT-4 Evaluation}
We adopt the template in Table~\ref{tab:gpt4-eval} to query GPT-4 and obtain the evaluation results with FairEval~\footnote{\url{https://github.com/i-Eval/FairEval}} to obtain more stable results.
Specifically, each tested instance is a quaternion: \texttt{(question, reference, response1, response2)}, where \texttt{response1} and \texttt{response2} are two responses from our Ying-VLM and the baseline model, respectively. 
For each instance, 
we query GPT-4 to judge which response is of better quality regarding accuracy, relevance and naturalness.
We populate the quaternion into the evaluation template to form two query prompts:
\texttt{T(Q=question, R=reference, R1=response1, R2=response2)} and \texttt{T(Q=question, R=reference, R1=response2, R2=response1)}.
We set the temperature of GPT-4 to $1$ and sample three completions for each query prompt.
Therefore, each response will receive $6$ scores, and we use the average score as the final score for each response.
The response with the higher final score is considered the better response.
The GPT-4 evaluation incurred a cost of  \$$20.45$ for InstructBlip and \$$20.90$ for MiniGPT-4.

\begin{table}[ht!]
\caption{Template used to query GPT-4 for evaluating the response quality of different models.}

\begin{tcolorbox}

[Question] 

\textcolor[rgb]{0,0,0.9}{\{Q\}}

[The Start of Reference Answer] 

\textcolor[rgb]{0,0,0.9}{\{R\}}

[The End of Reference Answer] 

[The Start of Assistant 1's Answer] 

\textcolor[rgb]{0,0,0.9}{\{R1\}}

[The End of Assistant 1's Answer]

[The Start of Assistant 2's Answer] 

\textcolor[rgb]{0,0,0.9}{\{R2\}}

[The End of Assistant 2's Answer] 

\ 

[System]

We would like to request your feedback on the performance of two AI assistants in response to the user's multimodal question displayed above.
We provided no multimodal inputs other than question text, but we provided a reference answer for this question. You need to evaluate the quality of the two responses based on the question and the reference answer.

Please rate the on the follow aspects:

1. Accuracy: whether the candidate's response is consistent with the original answer, this is important as we do not want a misleading result;

2. Relevance: whether the candidate's  response is highly relevant to the question and image content;

3. Naturalness: whether the candidate's response is engaging, providing a great communication experience for the user when interacting with the AI visual assistant. 

of the two Assistants' responses.

\ 

Each assistant receives an overall score on a scale of 1 to 10, where a higher score indicates better overall performance.

Please first provide a comprehensive explanation of your evaluation, avoiding any potential bias and ensuring that the order in which the responses were presented does not affect your judgment. 

Then, output two lines indicating the scores for Assistant 1 and 2, respectively.

\ 

Output with the following format:

Evaluation evidence: <evaluation explanation here>

The score of Assistant 1: <score>

The score of Assistant 2: <score>

\end{tcolorbox}

\label{tab:gpt4-eval}
\end{table}

\clearpage

\bibliographystyle{abbrvnat}
\bibliography{ref}
\end{document}